\documentclass[10pt,twocolumn,letterpaper]{article}
\usepackage[accsupp]{axessibility}
\usepackage{style}
\usepackage[dvipsnames, svgnames, x11names]{}
\usepackage{times}
\usepackage{epsfig}
\usepackage{graphicx}
\usepackage{subcaption}
\usepackage{amsmath}
\usepackage{amssymb}
\usepackage{adjustbox}
\usepackage{bm}
\usepackage[ruled,vlined]{algorithm2e}
\usepackage{multirow}
\usepackage{booktabs}
\usepackage{bbding}
\usepackage{multirow,booktabs}
\usepackage{colortbl}
\PassOptionsToPackage{hyphens}{url}
\usepackage[pagebackref=true,breaklinks=true,letterpaper=true,colorlinks,bookmarks=false]{hyperref}
\usepackage[capitalize]{cleveref}
\usepackage{float}

\newcommand{\PAR}[1]{\vskip4pt \noindent{\bf #1~}} 

\usepackage{multirow}

\begin{document}

\title{MotionStreamer: Streaming Motion Generation via Diffusion-based Autoregressive Model in Causal Latent Space
\vspace{-1em}}

\author{
Lixing Xiao$^{1}$ \ \ \ \ Shunlin Lu$^{2}$ \ \ \ \ Huaijin Pi$^{3}$ \ \ \ \ Ke Fan$^{4}$ \ \ \ \ Liang Pan$^{3}$ \\ Yueer Zhou$^{1}$ \ \ \ \ Ziyong Feng$^{5}$ \ \ \ \ Xiaowei Zhou$^{1}$ \ \ \ \ Sida Peng$^{1^{\dag}}$ \ \ \ \ Jingbo Wang$^{6}$ \\[0.7em]
$^{1}$Zhejiang University \ \ \ \ \ \
$^{2}$The Chinese University of Hong Kong (Shenzhen) \\
$^{3}$The University of Hong Kong \ \ \ \
$^{4}$Shanghai Jiao Tong University \\ 
\ \ \ \ \ \ \ \ \ $^{5}$DeepGlint \ \ \ \ \ \ \ \ \ \ \ \ \ \ \ \ \ \ \ \ \ \ \ \ \ \ 
$^{6}$Shanghai AI Laboratory \\
\texttt{lixingxiao0@gmail.com, pengsida@zju.edu.cn}
}

\twocolumn[{%
\renewcommand\twocolumn[1][]{#1}%
\maketitle
\begin{center}
    \centering
    \captionsetup{type=figure}
    \vspace{-6mm}
    \includegraphics[width=\textwidth]{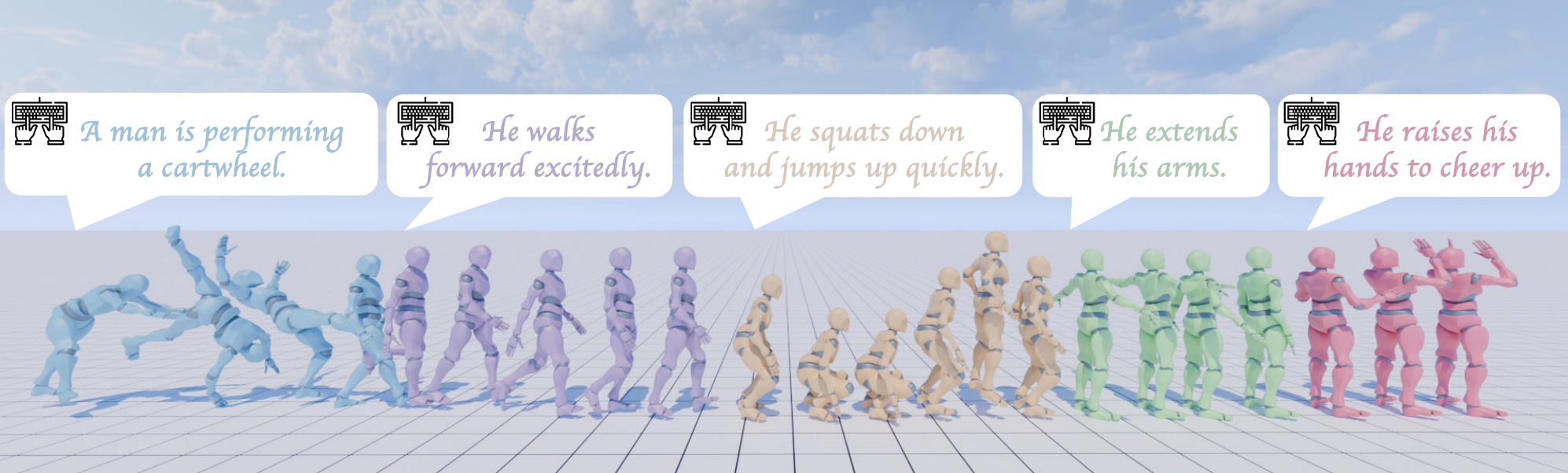}
    \captionof{figure}{\textbf{Visualization of streaming motion generation process}. Texts are incrementally inputted and motions are generated online.}
    \label{fig:teaser}
\end{center}%
}]
\let\thefootnote\relax\footnotetext{$^\dagger$ Corresponding author
}

\begin{abstract}
This paper addresses the challenge of text-conditioned streaming motion generation, which requires us to predict the next-step human pose based on variable-length historical motions and incoming texts.
Existing methods struggle to achieve streaming motion generation, e.g.,
diffusion models are constrained by pre-defined motion lengths, while GPT-based methods suffer from delayed response and error accumulation problem due to discretized non-causal tokenization.
To solve these problems, we propose MotionStreamer, a novel framework that incorporates a continuous causal latent space into a probabilistic autoregressive model.
The continuous latents mitigate information loss caused by discretization and effectively reduce error accumulation during long-term autoregressive generation.
In addition, by establishing temporal causal dependencies between current and historical motion latents, our model fully utilizes the available information to achieve accurate online motion decoding.
Experiments show that our method outperforms existing approaches
while offering more applications, including multi-round generation, long-term generation, and dynamic motion composition.
Project Page: \url{https://zju3dv.github.io/MotionStreamer/}
\end{abstract}

\section{Introduction}
\label{sec:intro}
Streaming motion generation aims to incrementally synthesizing human motions while dynamically adapting to online text inputs and maintaining semantic coherence.
Generating realistic and diverse human motions in a streaming manner is essential for various real-time applications, such as video games, animation, and robotics.
Streaming motion generation presents a significant challenge due to two fundamental requirements.
Firstly, the framework must incrementally process sequentially arriving textual inputs while maintaining online response.
Secondly, the model should be able to continuously synthesize motion sequences that exhibit contextual consistency by effectively integrating historical information with incoming textual conditions, ensuring alignment between progressive text semantics and kinematic continuity across extended timelines.

Conventional diffusion-based motion generation models \cite{MDM, MLD, FlowMDM} are constrained by their non-incremental generation paradigm with static text conditioning and fixed-length generation processes.
This inherently limits their ability to dynamically evolving textual inputs in streaming scenarios.
Other autoregressive motion generation frameworks \cite{T2M-GPT, mGPT} are able to generate motions in a streaming manner.
However, they have difficulties in achieving online response due to their non-causal tokenization architecture, which prevents partial token decoding until all sequence is generated.
Real-time motion generation methods like DART~\cite{Dart} face a critical limitation in their reliance on fixed-window local motion primitives,
which inherently restricts their capacity to model variable-length historical contexts and dynamically align with evolving textual inputs.

In this work, we propose a novel framework for streaming motion generation, named \textbf{MotionStreamer}.
The visualization of the streaming generation process is illustrated in Fig.~\ref{fig:teaser}.
Our core innovation is incorporating a diffusion head into an autoregressive model to predict the next motion latent, while introducing a causal motion compressor to enable online decoding in a streaming manner.
Specifically, given an input text, we extract the textual feature, combine it with historical motion latents, and use an autoregressive model to generate a condition feature, which guides a diffusion model to generate the next motion latent.
In contrast to previous methods~\cite{T2M-GPT, mGPT} that use vector quantization (VQ) based motion tokenizer and GPT architecture to generate discrete motion tokens, our continuous motion latents can avoid information loss of discrete tokens and accumulation of error during the streaming generation process, as demonstrated by our experimental results in Sec.~\ref{sec:quantitative_results}.

A causal temporal AutoEncoder is then employed to convert motion latent into the next human pose.
The causal network effectively establishes temporally causal dependencies between current and historical motion latents, allowing for online motion decoding.
The key to achieving streaming generation is enabling the model to dynamically extract relevant information from a variable-length history to guide the next motion prediction. 
To enable the autoregressive model to self-terminate without a pre-defined sequence length, we additionally encode an ``impossible pose" to get a reference end latent as the continuous stopping condition.

During experiments, we found that naive training of our model still suffers from error accumulation and cannot well support multi-round text input.
To address these issues, we propose two training strategies: Two-forward training and Mixed training.
Two-forward training strategy first generates motion latents using ground-truth,
then replaces partial ground-truth latents with first-forward predictions for a hybrid second-forward,
effectively mitigating the exposure bias inherent in autoregressive training while preserving parallel efficiency.
Mixed training strategy unifies atomic (text, motion) pairs and contextual (text, history motion, current motion) triplets in a single framework,
enabling compositional semantics learning and generalization to unseen motion combinations.

We evaluate our approach on the HumanML3D \cite{T2M} and BABEL \cite{BABEL} datasets, which are widely-used for text-to-motion benchmarks.
Across these datasets, our method achieves state-of-the-art performance on both text-to-motion and long-term motion synthesis tasks.
We also demonstrate the superiority of our method on abundant applications.
Such streaming generation framework is suitable for online multi-round generation with progressive text inputs,
long-term motion generation with multiple texts provided
and dynamic motion composition where subsequent motions can be regenerated
by altering textual conditions while preserving the initially generated prefix motion.

Overall, our contributions can be summarized as follows:
\begin{itemize}
    \item We propose MotionStreamer, a novel framework combining a diffusion head
    with an autoregressive model to directly predict continuous motion latents,
    which enables streaming motion generation with incremental text inputs.
\end{itemize}

\begin{itemize}
    \item We propose a causal motion compressor (Causal TAE) for continuous motion compression,
    which eliminates information loss from discrete quantization and
    establishes temporally causal latent dependencies to support streaming decoding and online response.
    We adopt Two-Forward training strategy to mitigate error accumulation in streaming generation scenarios.
  \end{itemize}

\begin{itemize}
    \item We demonstrate great performances of our approach on benchmark datasets.
    We also show various downstream applications,
    including online multi-round generation, long-term generation and dynamic motion composition.
\end{itemize}

\begin{figure*}[ht]
    \centering
    \resizebox{1\linewidth}{!}{
     \includegraphics[width=1\linewidth]{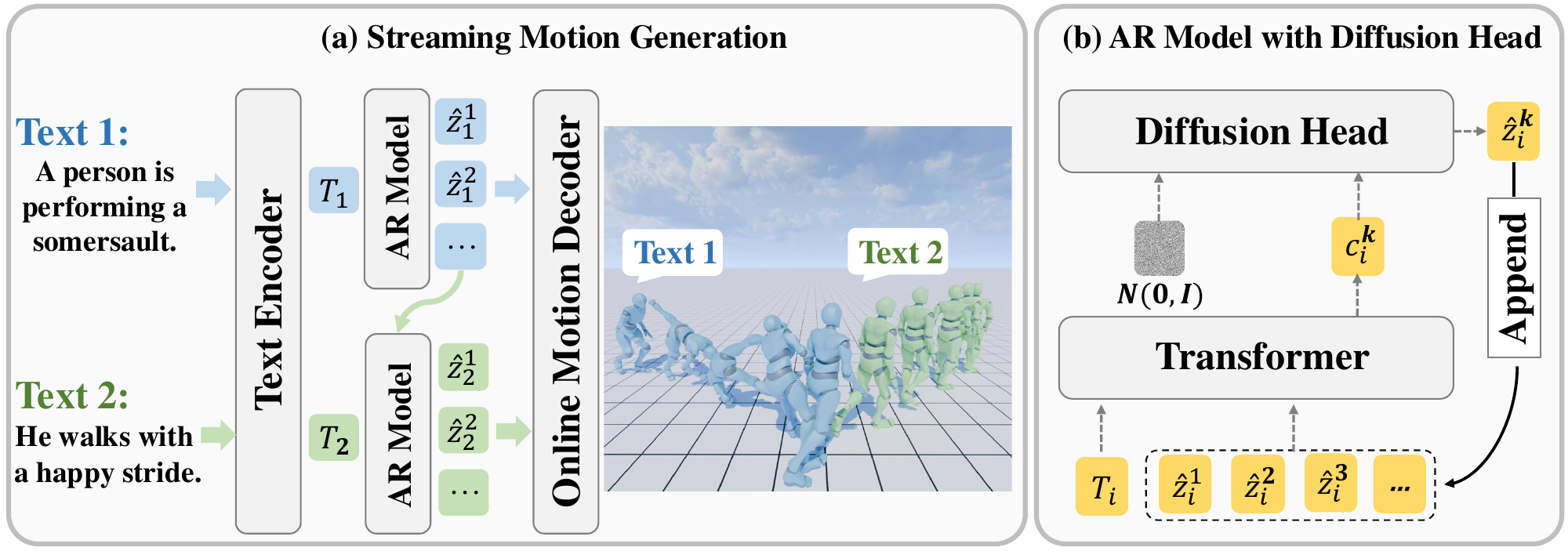}
     }
     \caption{\textbf{Overview of MotionStreamer.}
     During inference,
     the AR model streamingly predicts next motion latents conditioned on the current text and previous motion latents. Each latent can be decoded into motion frames online as soon as it is generated.
     }
     \label{fig:pipeline}
\end{figure*}

\section{Related Work}
\PAR{Text-conditioned motion generation.}
Text-conditioned motion generation aims to synthesize 3D human motions from natural language descriptions \cite{T2M}.
Various approaches \cite{pan2025tokenhsi, yang2025sigman, li2024dispose, chen2025dancetogether, yu2025hero, yu2025skillmimic, ji2025pomp, pi2025coda, meng2025absolute, pi2024motion, chen2023humanmac, pi2023hierarchical, wang2025timotion, xu2025mospa, zhang2025humanmm, chen2024motionllm, ji2023stylevr} have explored diverse architectures and methodologies to enhance the naturalness and expressiveness for digital human and motion animation.
Previous works \cite{T2M, TEMOS, TEACH} leverage VAE \cite{vae} to learn cross-modal mappings between text and motion spaces.
Some works \cite{MDM, MLD, motionlcm, motiondiffuse, FlowMDM, DoubleTake, PCMDM, multidiffusion, M2DM, freemotion, fg-t2m, sport} also apply diffusion models \cite{DDPM} to this task.
Another line of works \cite{T2M-GPT, mGPT, motion, AvatarGPT, attt2m, humantomato, scamo, fan2025go} first discretize motions into discrete tokens and employ autoregressive models (e.g., GPT) for sequential token prediction.
Furthermore, some approaches \cite{mmask, mmmodel, BAMM, fan2024textual, mardm} adopt a BERT-style \cite{BERT} bidirectional Transformer architecture \cite{maskgit} to reconstruct masked motion segments under text guidance.
However, most existing works only focus on offline generation, where the entire motion sequence is generated at once.
More recently, CAMDM \cite{camdm} and AMDM \cite{amdm} apply diffusion models into an autoregerssive manner for real-time interactive character control.
Ready-to-React \cite{ready_to_react} further explores this idea in two-character interaction.
CLoSD \cite{closd} and DART \cite{Dart} apply it for real-time text driven motion control.
However, these methods are not strictly causal,
as they rely on a fixed-length context window while our method could handle variable-length historical information
and incrementally generate motions in a streaming manner.

\PAR{Motion compression.}
Following the success of image generation \cite{stablediffusion}, previous works \cite{MLD, T2M-GPT} first encode the raw motion sequences into a latent space and then generate motions within it.
The most popular method is to use Vector Quantized Variational AutoEncoders (VQ-VAE) \cite{VQVAE} for motion tokenization.
TM2T \cite{TM2T} first introduces vector quantization for motion discretization.
T2M-GPT \cite{T2M-GPT} employs VQ-VAE to compress motion sequences into a discrete latent space and then uses a GPT for motion generation.
MoMask \cite{mmask} leverages Residual VQ-VAE (RVQ-VAE) \cite{rvqvae} to progressively reduce quantization errors.
In contrast, MLD \cite{MLD} utilizes standard VAEs \cite{vae} to convert a motion sequence into an embedding and then use a diffusion model to generate the latent.
However, existing methods require a whole motion sequence to be encoded and decoded, which is not suitable for streaming generation.
In this paper, we propose a causal motion compression approach to achieve streaming motion generation with online response.
\section{Method}
We address the task of streaming motion generation with online response by introducing a novel framework, named MotionStreamer.
The overview of the proposed framework is illustrated in Fig.~\ref{fig:pipeline}.
In section~\ref{sec:problem_formulation}, we first introduce the problem formulation of streaming motion generation and the motion representation used in this work.
In section~\ref{sec:causal_tae}, we introduce a Causal Temporal AutoEncoder for continuous motion compression and online decoding.
In section~\ref{sec:motionstreamer}, we present a diffusion-based autoregressive motion generator and the streaming generation process.

\subsection{Problem Formulation}
\label{sec:problem_formulation}
\PAR{Task definition.}
We first introduce the formulation of streaming motion generation.
In contrast to previous text-to-motion generation \cite{T2M} which is conditioned on a pre-defined fixed text prompt, we consider the case where a series of text prompts are given sequentially.
Given a streaming sequence of text prompts $\{\mathcal{P}_i\}_{i=1}^{M}$, the goal is to generate a sequence of motion frames $\{x_j\}_{j=1}^{N}$ online,
where $\mathcal{P}_i$ is the $i$-th text prompt and $x_j$ is the $j$-th frame pose.

\PAR{Motion Representation.}
Previous works \cite{T2M, mGPT, motion, mmask,MDM} mainly uses the 263-dimensional pose representation \cite{T2M} for motion generation.
However, this representation requires an additional post-processing step \cite{Smplify},
which is time-consuming and introduces rotation error \cite{humanmlissue} to be converted to SMPL \cite{SMPL} body parameters.
To overcome this issue, we slightly modify it and directly use SMPL-based 6D rotation \cite{SMPL} as joint rotations.
Similar to prior works on character control \cite{starke2019neural, shi2024interactive}, each pose \(x\) is represented by a 272-dimensional vector:
\begin{equation}
x = \{\dot{r}^x,\dot{r}^z,\dot{r}^a,\,j^p,\,j^v,\,j^r\}.
\end{equation}
where we project the root on the XZ-plane (ground plane), \((\dot{r}^x, \dot{r}^z \in \mathbb{R})\) are root linear velocities on the XZ-plane,
\(\dot{r}^a \in \mathbb{R}^6\) denotes root angular velocity represented in 6D rotations,
\(j^p \in \mathbb{R}^{3K}\), \(j^v \in \mathbb{R}^{3K}\), and \(j^r \in \mathbb{R}^{6K}\) are local joint positions, local velocities, and local rotations
relative to the root space, \(K\) is the number of joints.
For SMPL \cite{SMPL} character, $K=22$, then we get the $2 + 6 + 3\times22 + 3\times22 +6\times22 = 272$ dimensions.
This representation removes the post-processing step and we could directly use it for animating a SMPL character model.

\subsection{Causal Temporal AutoEncoder}
\label{sec:causal_tae}
Streaming motion generation requires online motion decoding for dynamic text inputs.
However, most existing works \cite{T2M-GPT, mGPT, motion, mmask} utilize temporal VQ-VAE to decode the whole sequence at once, where each frame inherently depends on past and future frames.
Furthremore, the reliance on discrete tokenization induces quantization error accumulation across tokens, progressively degrading motion coherence in streaming generation scenarios.
To address these issues, we introduce a Causal Temporal AutoEncoder (Causal TAE) to enable motion generation in a causal latent space.

\begin{figure}
    \centering
     \includegraphics[width=1\linewidth]{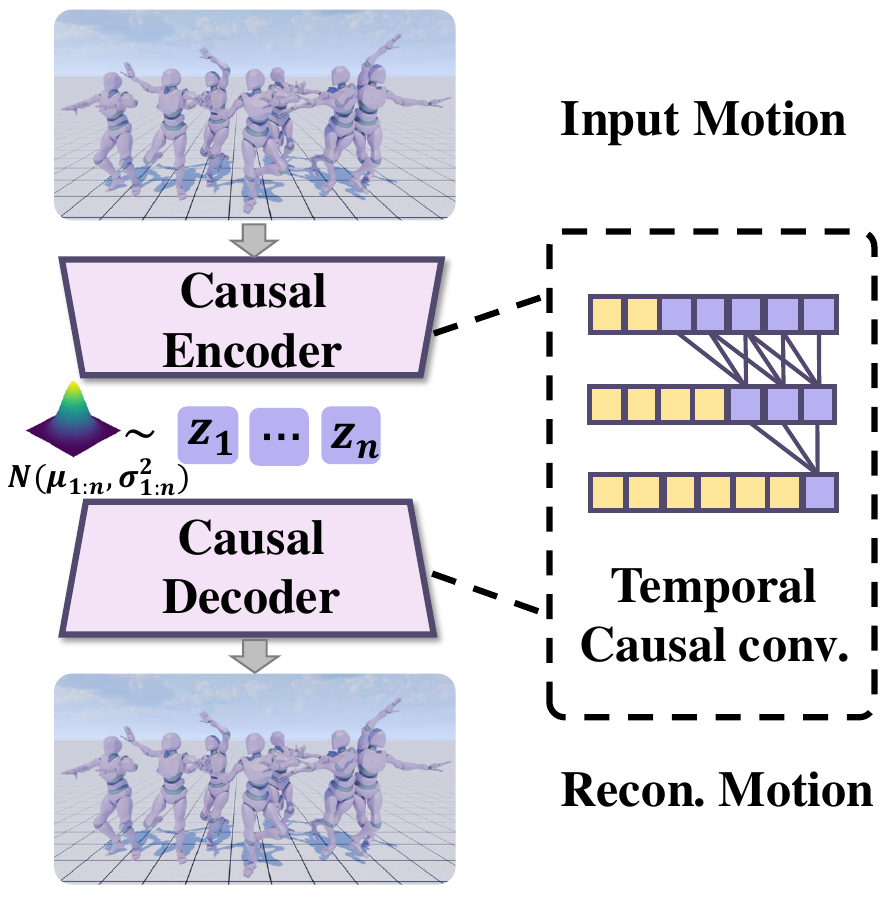}
     \caption{\textbf{Architecture of Causal TAE.}
     1D temporal causal convolution is applied in both the encoder and decoder.
     Variables $z_{1:n}$ are sampled as continuous motion latent representations.
     }
     \label{fig:recon}
\end{figure}

\PAR{Architecture.}
Causal TAE is designed to achieve continuous motion compression while explicitly modeling temporal dependencies and enforcing causal constraints for sequential motion representation.
Fig.~\ref{fig:recon} shows the proposed Causal TAE network.
We employ 1D causal convolution \cite{Magvitv2} for constructing temporal encoder $\mathcal{E}$ and decoder $\mathcal{D}$ to convert raw motion sequences into a causal latent space.
The causality is guaranteed by a temporal padding scheme.
Specifically, for a convolution layer with kernel size $k_t$, stride $s_t$ and dilation rate $d_t$,
we pad $(k_t-1) \times d_t + (1-s_t)$ frames at the beginning of the sequence.
In this way, each frame only depends on the frames before it and the future frames are not involved in the computation.
Moreover, explicitly modeling temporal causal structures in the latent space enables the model to
learn temporal and causal dependencies inherent in the causally-related motion data.

Given a motion sequence $X = \{x_1, x_2, \cdots, x_N\}$ with $x_t \in \mathbb{R}^{D}$, 
where $N$ is the number of frames and $D$ is the motion dimension ($D=272$),
we could obtain a set of temporal Gaussian distribution parameters $\{\mu_{1:N/l}, \sigma^2_{1:N/l}\}$ and preform reparameterization \cite{vae} to get continuous motion latent representation $Z = \{z_{1}, z_{2}, \cdots, z_{N/l}\}$ with $z_{i} \in \mathbb{R}^{d_{c}}$,
$l$ represents the temporal downsampling rate of the Encoder $\mathcal{E}$.
This architecture reconstructs motion frames while strictly preserving temporal causality across the sequence.

\PAR{Training Objective.}
We use the same loss function as $\sigma$-VAE \cite{sigma-vae} to train the Causal TAE.
In order to further enhance the reconstruction stability of the root joint, we add a root joint loss $\mathcal{L}_{root}$.
The full loss function is defined as:
\begin{equation}
    \label{eq:recon_loss}
    \mathcal{L} = \mathcal{L}_{recon} + D_{KL}(q(z|x)||p(z)) + \lambda \mathcal{L}_{root}.
\end{equation}
where 
\begin{equation}
    \label{eq:tae_loss}
    \mathcal{L}_{recon} = \sum_{d=1}^{D}\sum_{i=1}^{N}(\frac{(x_{di}-\hat{x}_{di})^2}{2\sigma^{*^{2}}}+ln\sigma^{*}),
\end{equation}
\begin{equation}
    \label{eq:root_loss}
    \mathcal{L}_{root} = \sum_{d=1}^{D_{root}}\sum_{i=1}^{N}(\frac{(x_{di}-\hat{x}_{di})^2}{2\sigma^{*^{2}}}+ln\sigma^{*}),
\end{equation}

\setlength{\abovedisplayskip}{-30pt}
\begin{equation}
    \label{eq:tae_loss}
    \sigma^{*^{2}} = MSE(x, \hat{x})
                   = \frac{1}{DN}\sum_{d=1}^{D}\sum_{i=1}^{N}(x_{di}-\hat{x}_{di})^2,
\end{equation}

\setlength{\abovedisplayskip}{-7pt} 
\begin{equation}
    \label{eq:tae_loss}
    D_{KL}(q(z|x)||p(z)) = \frac{1}{2}\sum_{d=1}^{d_{c}}\sum_{i'=1}^{N/l}(\mu_{di'}^2+\sigma_{di'}^2-ln(\sigma_{di'}^2)-1).
\end{equation}

$D$, $D_{root}$ and $d_{c}$ represent the dimensions of the motion, root joint and the latent representation respectively.
Specifically, $D=272$, $D_{root}=8$ (the first 8 dims relate to the root joint). $d_{c}=16$ is the best choice in our experiments.
$N$ and $l$ represent the number of frames and temporal downsampling rate, $l$ is set to $4$ here.
$x_{di}$ and $\hat{x}_{di}$ are the ground-truth motion and the reconstructed motion at the $i$-th frame of the $d$-th dimension,
$\sigma^{*}$ is the analytic solution of the standard deviation \cite{sigma-vae}.
$D_{KL}$ represents the KL divergence, $q(z|x)$ is the distribution of latents given the motions,
$p(z) = \mathcal{N}(0, I)$ is the prior distribution.
($\mu_{di'}$, $\sigma_{di'}^2$) are the Gaussian distribution parameters at the $i'$-th latent of the $d$-th latent dimension, which derive from the Causal TAE Encoder $\mathcal{E}$.
$\lambda$ is the balancing hyperparameter.

Causal TAE offers distinct technical advantages for motion compression.
Its causal property inherently supports online decoding without requiring access to future frames, which is critical for streaming generation.
With the employment of continuous token representation, it bypasses the discretization bottleneck of existing VQ-based methods.

\subsection{MotionStreamer}
\label{sec:motionstreamer}
In this section, we present MotionStreamer, a streaming generation pipeline based on a causally-structured latent space.
In order to handle coherence between arriving text inputs and historical generated motions, we hypothesize that current motion should only be conditioned on previous motion and current text.
As illustrated in Fig.~\ref{fig:pipeline}, MotionStreamer comprises a pre-trained text encoder, a diffusion-based autoregressive model, and the online motion decoder (the learned Causal TAE decoder).

\PAR{Training.}
Each training sample can be represented as: $S_i = (T_i, C_i, Z_i)$,
where $T_i \in \mathbb{R}^{1 \times d_{t}}$ is the text embedding obtained via a pre-trained language model (e.g. T5-XXL \cite{T5}),
$C_i \in \mathbb{R}^{k \times d_{c}}$ and
$Z_i \in \mathbb{R}^{n \times d_{c}}$ are the previous motion latents and current motion latents encoded by the learned Causal TAE,
where $k$, $n$, $d_t$ and $d_{c}$ denote the lengths of previous motion latents, current motion latents,
the text embedding dimension and the latent dimension respectively.
We concatenate them along temporal axis to form a sequence $S_i = [T_i, C_i, Z_i]$.
We employ a diffusion-based autoregressive Transformer to predict motion latents.
The latent sequence $S$ is first processed by the Transformer and a causal mask is applied to ensure the temporal causality \cite{T2M-GPT}.
After the Transformer processing, we obtain the intermediate latents $\{c_{i}^{1}, c_{i}^{2}, \cdots, c_{i}^{n}\}$, which serve as the condition for
the diffusion head (a small MLP) to predict motion latents $\{\hat{z}_{i}^{1}, \hat{z}_{i}^{2}, \cdots, End_{i}\}$.
$End_{i}$ is the reference end latent inserted at the end of a sequence as the continuous stopping condition, which we will elaborate later.
Following \cite{MAR, DDPM}, the loss function is defined as:
\setlength{\abovedisplayskip}{7pt}
\begin{equation}
    \label{eq:diffusion_loss}
    \mathcal{L} = \mathbb{E}_{\epsilon,t} [||\epsilon-\epsilon_{\theta}(Z_t|t,C_{i},T_{i})||^2].
\end{equation}
where $t$ denotes the timestep of noise schedule.
We employ QK normalization (i.e., normalize both queries and keys) \cite{qknorm} before self-attention layer to enhance training stability.

\noindent\textbf{Two-Forward strategy.}
We observe that using teacher-forcing \cite{exposure} directly during training often leads to error accumulation in the autoregressive generation process.
To this end, we propose a Two-Forward strategy that progressively introduces the test-time distribution during training. 
Specifically, after the first forward pass, we replace a subset of ground-truth motion latents with their generated counterparts, creating a mixture of real and generated motion latents. 
This hybrid input is then used in the second forward pass, where gradients are backpropagated. 
We employ a cosine scheduler to control the proportion of replaced motion latents. More details are provided in Sec.A of the appendix.

\PAR{Mixed training.}
The datasets contain two types of training samples, so we set $C_i$ to Null if there is no historical motion in the dataset.
We find that this simple strategy enables a seamless transition between two consecutive motions.

\PAR{Continuous stopping condition.}
\label{sec:stopping_condition}
Streaming generation requires automatically determining the generation length for each text prompt.
Previous method \cite{meng2024autoregressive} uses a binary classifier to determine whether to stop generation, which suffers from a strong class imbalance.
In contrast, we introduce an ``impossible pose" prior (i.e., all-zero vectors $\mathbf{0} \in \mathbb{R}^{D}$) as the stopping condition and use the causal TAE to convert it into the latent space.
The encoded latent serves as the reference end latent.
The generation should stop when the distance between the currently generated latent and the reference end latent is less than a threshold.
Therefore, MotionStreamer is able to stop generation automatically and enables online and multi-round generation.

\PAR{Inference.}
During inference, given a stream of text prompts $\{\mathcal{P}_{i}\}_{i=1}^{M}$,
the first text embedding $T_{1}$ is first fed into the autoregressive motion generator to generate the first predicted motion latent sequence $\hat{Z}_{1}=\{\hat{z}_{1}^{1}, \hat{z}_{1}^{2}, \cdots, \hat{z}_{1}^{n_{1}}\}$.
As soon as a motion latent is predicted, it can be immediately processed by the online motion decoder (i.e., the learned Causal TAE decoder) to get the output motion frames, benefiting from its causal property.
If the distance between the currently predicted motion latent and the reference end latent is lower than a threshold, the generation process of this prompt stops.
Then, we replace $T_{1}$ with $T_{2}$ as the current text embedding.
The already generated motion latent sequence $\hat{Z}_{1}$
is appended to the end of the second text embedding, forming the contextual latents used as input for the next autoregressive step.
We then generate the second predicted motion latent sequence $\hat{Z}_{2}=\{\hat{z}_{2}^{1}, \hat{z}_{2}^{2}, \cdots, \hat{z}_{2}^{n_{2}}\}$.
Next, we replace $T_{2}$ with future text embedding, removes $\hat{Z}_{1}$ from the condition latents and uses $\hat{Z}_{2}$ as the historical motion latents.
Therefore, the third sequence could be predicted.
This streaming generation process is repeated until the entire motion sequence $\{\hat{Z}_{i}\}_{i=1}^{N}$ is generated, ensuring online response during streaming generation process.

\section{Experiment}
\subsection{Experimental Setup}
\textbf{Dataset.}
We evaluate the proposed MotionStreamer on HumanML3D \cite{T2M}
and BABEL \cite{BABEL} datasets, with the original train and test splits.
The HumanML3D dataset integrates motion sequences
with three distinct textual descriptions.
The BABEL dataset provides frame-level textual descriptions with explicit inter-segment transition labels.
Unlike recent methods \cite{DoubleTake, FlowMDM} that use different motion representations for both HumanML3D and BABEL datasets,
we employ the 272-dimensional motion representation as mentioned in Sec.~\ref{sec:problem_formulation} for both datasets.
All motion sequences are uniformly resampled to 30 FPS.

\noindent \textbf{Evaluation Metrics.}
We adopt the metrics from \cite{T2M} for evaluation,
including: (1) Frechet Inception Distance (FID) \cite{FID}, indicating the distribution distance between the generated and real motion;
(2) Mean Per Joint Position Error (MPJPE), the average distance between the predicted and ground-truth joint positions, measuring the reconstruction quality;
(3) R-Precision (Top-1, Top-2, and Top-3 accuracy), the accuracy of the top-k retrieved motions;
(4) Multimodal Distance (MM-Dist), the average Euclidean distances between the generated motion feature and its text feature.
(5) Diversity, the average Euclidean distances of the randomly sampled 300 motion pairs, measuring the diversity of motions.
(6) Peak Jerk (PJ) \cite{FlowMDM}, the maximum value throughout the transition motion over all joints.
(7) Area Under the Jerk (AUJ) \cite{FlowMDM}, the area under the jerk curve. Both PJ and AUJ measures the smoothness of motions.

\subsection{Quantitative Results}
\label{sec:quantitative_results}
\noindent \textbf{Comparison on Text-to-Motion Generation.}
We trained an evaluator based on TMR \cite{TMR} to evaluate the quality of the generated motions.
We use the processed 272-dimensional motion data from HumanML3D \cite{T2M} train set for text-to-motion model training.
All baselines are trained from scratch following their original implementations.
As shown in Tab.~\ref{tab:t2m_results}, our method receives better performance across multiple metrics on HumanML3D \cite{T2M} test set . 

\begin{table}[t]
    \centering
    \setlength{\tabcolsep}{2pt}
    \small
    \begin{tabular}{lcccccc}
    \toprule
    Methods & FID $\downarrow$ & R@1 $\uparrow$ & R@2 $\uparrow$ & R@3 $\uparrow$ & MM-D $\downarrow$ & Div $\rightarrow$ \\
    \midrule
    Real motion & 0.002 & 0.702 & 0.864 & 0.914 & 15.151 & 27.492 \\
    \midrule
    MDM \cite{MDM} & 23.454 & 0.523 & 0.692 & 0.764 & 17.423 & 26.325 \\
    MLD \cite{MLD} & 18.236 & 0.546 & 0.730 & 0.792 & 16.638 & 26.352 \\
    T2M-GPT \cite{T2M-GPT} & 12.475 & 0.606 & 0.774 & 0.838 & 16.812 & \underline{27.275} \\
    MotionGPT \cite{mGPT} & 14.375 & 0.456 & 0.598 & 0.628 & 17.892 & 27.114 \\
    MoMask \cite{mmask} & \underline{12.232} & \underline{0.621} & \underline{0.784} & \underline{0.846} & 16.138 & 27.127 \\
    AttT2M \cite{attt2m} & 15.428 & 0.592 & 0.765 & 0.834 & \textbf{15.726} & 26.674 \\
    Ours & \textbf{11.790} & \textbf{0.631} & \textbf{0.802} & \textbf{0.859} & \underline{16.081} & \textbf{27.284} \\
    \bottomrule
    \end{tabular}
    \caption{\textbf{Comparison with baseline text-to-motion generation methods} on HumanML3D \cite{T2M} test set.
    MM-D and Div denote Multimodal Distance and Diversity respectively.}
    \label{tab:t2m_results}
\end{table}

\noindent \textbf{Comparison on Long-Term Motion Generation.}
We adopt a Mix Training Strategy for streaming long-term generation training.
Specifically, we create training samples by pairing adjacent subsequences from long motion sequences in BABEL.
Additionally, we incorporate text-motion pairs from the HumanML3D dataset for mix training.
The comparison results on BABEL \cite{BABEL} dataset are demonstrated in Tab.~\ref{tab:babel_results}.
Following FlowMDM~\cite{FlowMDM}, the motion transition length is set to 30 frames.
We modified T2M-GPT to support streaming generation (marked as T2M-GPT*)
and also adapted our model to use VQ for discretization (marked as VQ-LLaMA).
Experimental results show that neither the existing long-term generation baseline nor the discrete autoregressive model performs as well as our streaming generation approach in the continuous latent space.

\begin{table*}
\centering
    \begin{tabular}{lcccccccc}
    \toprule
    \multicolumn{1}{c}{Methods} & \multicolumn{4}{c}{Subsequence} & \multicolumn{4}{c}{Transition} \\
    \cmidrule(lr){2-5} \cmidrule(lr){6-9}
      & R@3 $\uparrow{}$ & FID $\downarrow{}$ & Div $\rightarrow$ & MM-D $\downarrow{}$  & FID $\downarrow{}$ & Div $\rightarrow{}$ & PJ $\rightarrow{}$ & AUJ $\downarrow{}$  \\
    \midrule
    GT  & 0.634 & 0.000 & 24.907 & 17.543 & 0.000 & 21.472 & 0.03 & 0.00  \\
    \midrule
    DoubleTake \cite{DoubleTake} & 0.452 & 23.937 & 22.732 & 21.783 & 51.232 & 18.892 & 0.48 & 1.83 \\
    FlowMDM \cite{FlowMDM} & \underline{0.492} & \underline{18.736} & \underline{23.847} & \underline{20.253} & \underline{34.721} & \underline{20.293} & \underline{0.06} & \textbf{0.51} \\
    T2M-GPT* \cite{T2M-GPT} & 0.364 & 39.482 & \textbf{24.317} & 20.692 & 43.823 & \textbf{20.797} & 0.12 & 1.43 \\
    VQ-LLaMA  & 0.383 & 24.342 & 19.329 & 38.285 & 36.293 & 19.932 & 0.08 & 1.20 \\
    Ours & \textbf{0.568} & \textbf{15.743} & 23.546 & \textbf{15.397} & \textbf{32.888} & 19.986 & \textbf{0.04} & \underline{0.90} \\
    \bottomrule
    \end{tabular}
    \caption{\textbf{Comparison with long-term motion generation methods} on BABEL \cite{BABEL} dataset.}
    
    \label{tab:babel_results}
\end{table*}

\begin{figure}
    \centering
    \resizebox{1\linewidth}{!}{
     \includegraphics[width=1\linewidth]{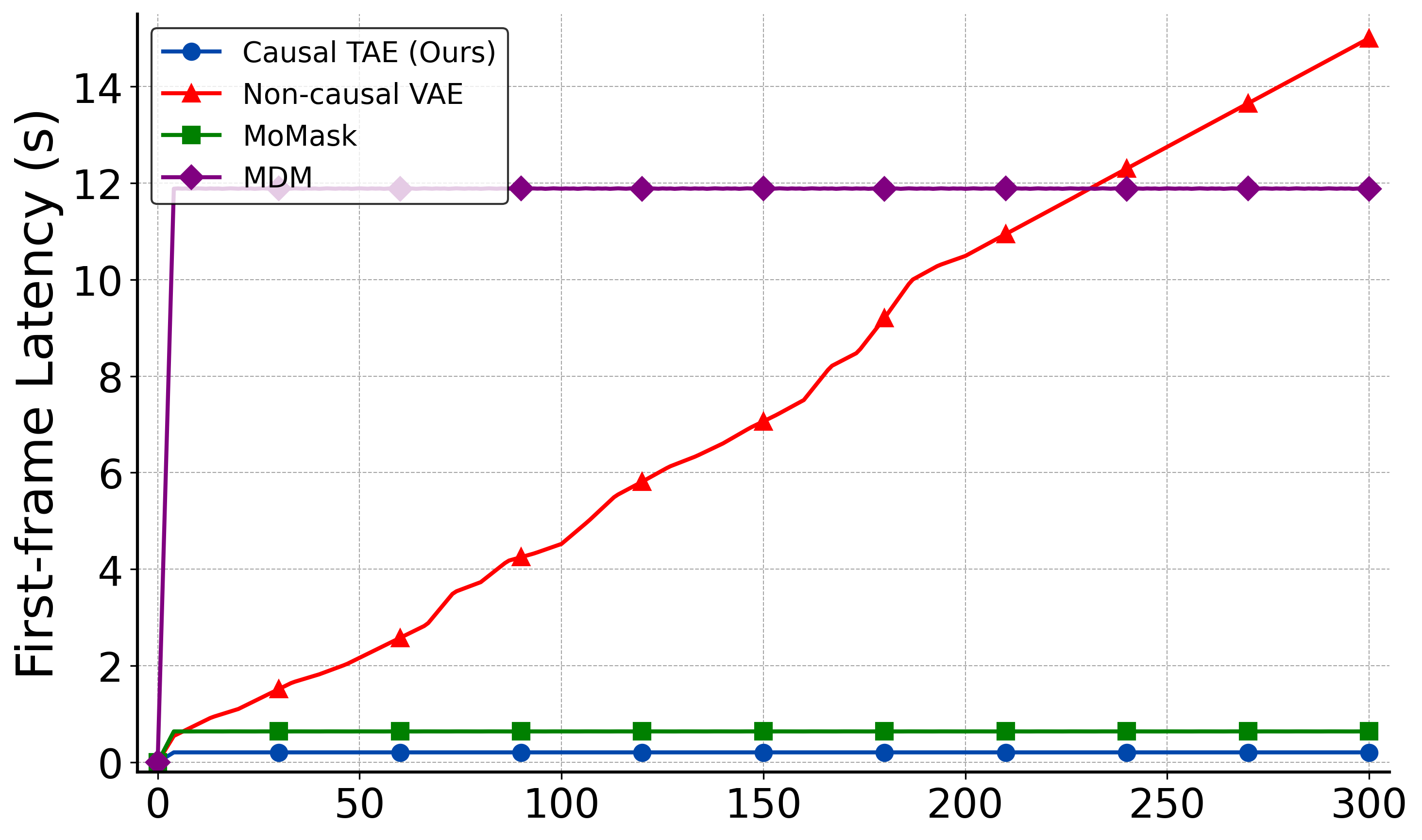}
     }
     \caption{\textbf{Comparison on the First-frame Latency} of different methods.
     Horizontal axis: the number of generated frames.
     Vertical axis: the time required to produce the first output frame.}
     \label{fig:time_comparison}
\end{figure}

\noindent \textbf{Comparison on First-frame Latency.}
As streaming generation requires the model to generate motion progressively and respond online.
Therefore, we adopt the First-frame Latency to evaluate the efficiency of different methods.
First-frame Latency refers to the time taken by the model to produce its first predicted frame,
serving as a key metric for evaluating online response ability.
Experimental results in Fig.~\ref{fig:time_comparison} show that our proposed Causal TAE achieves the lowest First-frame Latency,
benefiting from the causal property of motion latents, which can be decoded immediately after generation.
In contrast, non-causal VAE must wait until the entire sequence is generated before decoding,
causing First-frame Latency to increase as the number of generated frames grows.
Another fixed-length generation methods like \cite{MDM, mmask} exhibit higher First-frame Latency
as they process the entire sequence at once rather than generating frames progressively in a streaming manner.

\begin{figure*}[ht]
    \centering
    \resizebox{1\linewidth}{!}{
     \includegraphics[width=1\linewidth]{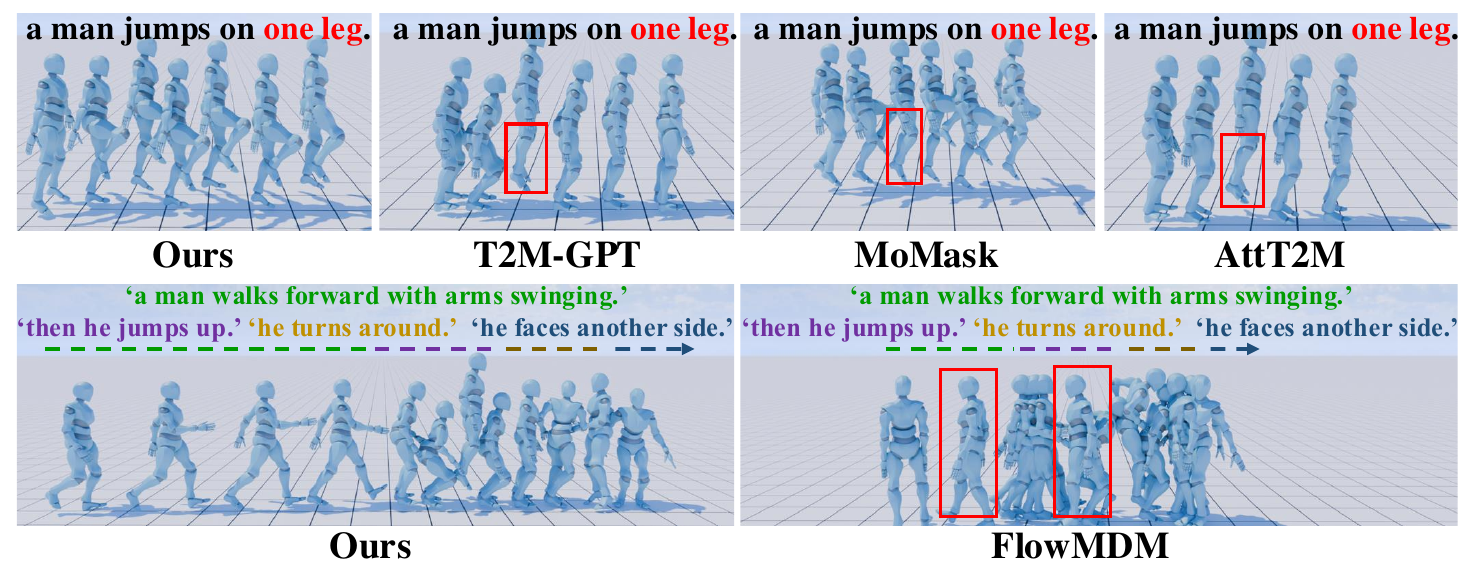}
     }
     \caption{\textbf{Visualization results} between our method and some baseline methods \cite{T2M-GPT, mmask, attt2m, FlowMDM}.
     The first row shows text-to-motion generation results, the second row shows long-term generation results
     and the third row shows the application of dynamic motion composition.
     }
     \label{fig:vis}
\end{figure*}

\subsection{Qualitative Results}
Figure~\ref{fig:vis} shows the qualitative results of our method compared with T2M-GPT \cite{T2M-GPT}, MoMask \cite{mmask}, AttT2M \cite{attt2m}, and FlowMDM \cite{FlowMDM}.
For the text-to-motion generation, we observe that VQ-based methods
have difficulty in generating motions that are accurate and aligned with the textual description.
In the case of ``a man jumps on one leg.", T2M-GPT \cite{T2M-GPT} and AttT2M \cite{attt2m} generate a motion where the person jumps with both legs instead. 
MoMask \cite{mmask} employs residual vector quantization (RVQ) to reduce quantization errors but still suffers from fine-grained motion details loss.
Specifically, the generated motion starts with a one-leg jump but later switches to two-leg jumps or alternating legs, along with noticeable sliding artifacts.
However, our method can generate motions that are more accurate with more details preserved as we use a continuous latent space without discretization process.

For long-term motion generation, we compare with FlowMDM \cite{FlowMDM} with a stream of prompts: [``a man walks forward with arms swinging.", ``then he jumps up.", ``he turns around.", ``he faces another side."].
The visualization results show that FlowMDM fails to generate the initial ``walking" motion,
instead producing in-place stepping.
However, we can generate more coherent and natural long-term motions streamingly as our model has the ability to dynamically extract relevant information from variable-length motion histories.
Please refer to the supplementary videos in our project page for more dynamic visualizations.

\begin{figure*}[ht]
    \centering
    \resizebox{1\linewidth}{!}{
     \includegraphics[width=1\linewidth]{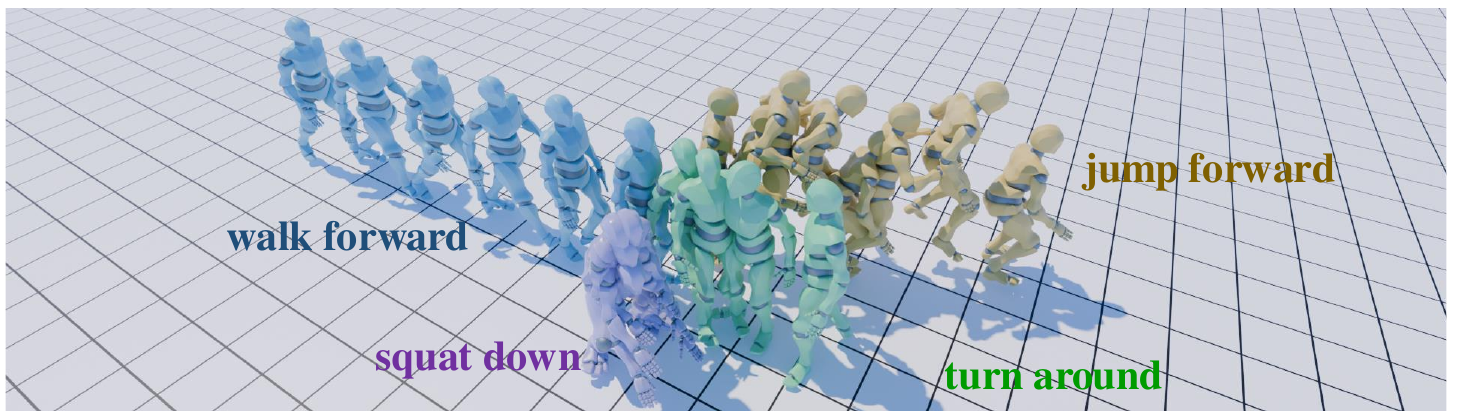}
     }
     \caption{\textbf{Dynamic motion composition.} Our model supports composition of multiple motions with different textual descriptions while
     maintaining previous motions unchanged.
     }
     \label{fig:application}
\end{figure*}

\subsection{Ablation Study}
\noindent \textbf{Architecture of the Causal TAE.}
We comprehensively evaluate the reconstruction performance and the corresponding generation quality
of different Causal TAE architectures on the HumanML3D \cite{T2M} test set, as shown in Tab.~\ref{tab:latent_ablation}.
We replace the motion compression stage with VQ-VAE \cite{T2M-GPT} to discretize the motions,
while keeping the second-stage model architecture identical to ours.
We also experimented with replacing Causal TAE with a non-causal temporal VAE and a standard temporal AE without vector quantization.
The results show that our continuous representation avoids the VQ process,
effectively reducing information loss and minimizing quantization error, thus performs better than
the VQ-VAE baseline.
The non-causal VAE performs worse than Causal TAE in both reconstruction and generation,
as Causal TAE inherently models the causal structure of motion data during compression.
This causal latent space is better suited for autoregressive generation, aligning naturally with the causal masking process.
While AE achieves the best reconstruction quality by learning a near-identity mapping,
its generation performance is significantly worse.
This highlights the crucial role of latent space representation in determining the effectiveness of subsequent motion generation.

We provide a more detailed ablation on the latent dimension and hidden size of Causal TAE,
as shown in Sec.B in the appendix.
Notably, we observe that
a larger latent dimension results in less compression rate, improving reconstruction quality.
However, this comes at the cost of poorer generation performance,
as insufficient compression and ineffective latent space representation makes it harder for the model to learn meaningful motion generation.
Meanwhile, the hidden size determines the model's capacity,
requiring a careful balance between compression rate and hidden size to ensure high reconstruction quality while enhancing generation performance.
Ablation on the hyperparameter $\lambda$ is provided in Sec.A of the appendix.

\noindent \textbf{Design choices of AR Model.}
We analyze the impact of different design choices of the AR model,
as shown in Tab.~\ref{tab:training_ablation}.
The results show that the QK normalization and Two-forward strategy are effective.
We also remove the diffusion head and use MSE loss for autoregressive training,
which leads to a significant drop in generation quality.
Moreover, we find that using T5-XXL \cite{T5} improves the generation performance compared with the CLIP \cite{clip} tokenizer.
We found that applying a binary classifier to predict whether to stop generation, as in \cite{meng2024autoregressive},
fails to learn the correct stopping condition.
As a result, we did not evaluate the model without the proposed continuous stopping condition.

\begin{table}
    \centering
    \setlength{\tabcolsep}{1pt}
    \small
    \begin{tabular}{lcccccc}
    \toprule
    \multirow{2}{*}{Methods} & \multicolumn{2}{c}{Reconstruction} & \multicolumn{4}{c}{Generation} \\
    
    \cmidrule(lr){2-3} \cmidrule(lr){4-7}
    ~ & FID $\downarrow$ & MPJPE $\downarrow$ & FID $\downarrow$ & R@3 $\uparrow$ & MM-D $\downarrow$ & Div $\rightarrow$ \\
    \midrule
    Real motion & - & - & 0.002 & 0.914 & 15.151 & 27.492 \\
    \midrule
    VQ-VAE & 5.173 & 63.9 & \underline{13.226} & \underline{0.824} & \underline{16.746} & 27.024 \\
    AE & \textbf{0.001} & \textbf{1.7} & 43.828 & 0.463 & 22.040 & \textbf{27.382} \\
    VAE & 2.092 & 26.2 & 19.902 & 0.735 & 17.926 & \underline{27.312} \\
    Ours & \underline{0.661} & \underline{22.9} & \textbf{11.790} & \textbf{0.859} & \textbf{16.081} & 27.284 \\
    \bottomrule
    \end{tabular}
    \caption{\textbf{Ablation Study of different motion compressors} on HumanML3D \cite{T2M} test set.
    MPJPE is measured in millimeters.}
    \label{tab:latent_ablation}
\end{table}

\subsection{Applications}
MotionStreamer offers various applications, including multi-round generation, long-term generation, and dynamic motion composition.
(1) \textbf{Multi-round generation} requires iteratively generating motion in response to sequential or interactive textual inputs.
Our model can process incremental text inputs, respond online, and autonomously determine when to stop generation.
(2) \textbf{Long-term generation} requires smoothly generating long motion sequences in response to sequential textual inputs.
(3) \textbf{Dynamic motion composition} refers to the capability of seamlessly integrating diverse motion sequences while preserving the consistency of previously generated motion prefix, as shown in Fig.~\ref{fig:application}.
Our Causal TAE enables this application,
which eliminates the need for full-sequence re-decoding
during the generation of subsequent motion latents,
requiring only the decoding of the newly generated motion latents.

\begin{table}[t]
    \centering
    \setlength{\tabcolsep}{2pt}
    \small
    \begin{tabular}{lcccccc}
    \toprule
    AR Design choices & FID $\downarrow$ & R@3 $\uparrow$ & MM-D $\downarrow$ & Div $\rightarrow$ \\
    \midrule
    Real motion & 0.002 & 0.914 & 15.151 & 27.492 \\
    \midrule
    w/o QK Norm & \underline{12.324} & 0.836 & 16.524 & 27.126 \\
    w/o Two-Forward & 12.944 & \underline{0.842} & \underline{16.432} & 27.282 \\
    w/o Diffusion Head & 59.296 & 0.360 & 22.864 & \textbf{27.325}\\
    CLIP & 14.234 & 0.791 & 17.534 & \underline{27.316} \\
    Ours & \textbf{11.790} & \textbf{0.859} & \textbf{16.081} & 27.284 \\
    \bottomrule
    \end{tabular}
    \caption{\textbf{Analysis of design choices of the AR model} on HumanML3D \cite{T2M} test set.
    CLIP indicates the use of CLIP model \cite{clip} as the text encoder to extract text features.}
    \label{tab:training_ablation}
\end{table}

\section{Conclusion}
\indent We present MotionStreamer, a novel framework for streaming motion generation that
integrates diffusion-based autoregressive model to directly predict causal motion latents.
By introducing Causal TAE,
MotionStreamer supports online response with progressive textual inputs.
We propose a Two-Forward training strategy to mitigate cumulative errors in streaming generation process.
Our method outperforms baseline approaches, demonstrating its competitiveness in motion generation while providing greater flexibility. It can be applied to multi-round motion generation, long-term motion generation, and dynamic motion composition.

\section*{Acknowledgement}
This work was partially supported by the National Key R\&D Program of China (No. 2024YFB2809102), NSFC (No. 62402427, NO. U24B20154), Zhejiang Provincial Natural Science Foundation of China (No. LR25F020003), DeepGlint, Zhejiang University Education Foundation Qizhen Scholar Foundation, and Information Technology Center and State Key Lab of CAD\&CG, Zhejiang University, 
the National Key R\&D Program of China (2022ZD0160201), and Shanghai Artificial Intelligence Laboratory.
{
    \small
    \bibliographystyle{ieeenat_fullname}
    \bibliography{main}
}

\clearpage
\renewcommand{\thesection}{\Alph{section}} 
\setcounter{section}{0}
\setcounter{page}{1}
\twocolumn[
\section*{\begin{center}\LARGE Appendix of ``MotionStreamer: Streaming Motion Generation via Diffusion-based Autoregressive Model in Causal Latent Space" \end{center}}
\vspace*{2em}
]
\renewcommand{\theHsection}{appendix.\Alph{section}}

\vspace{5em}
\section{Implementation Details}
\label{app:im_detail}
For the Causal TAE, both the encoder and decoder are based on the 1D causal ResNet blocks \cite{resnet}.
The temporal downsampling rate $l$ is set to 4 and all motion sequences are cropped to $N=64$ frames during training.
We train the first 1900K iterations with a learning rate of 5e-5 and the remaining 100K iterations with a learning rate of 2.5e-6.
We use the AdamW optimizer \cite{AdamW} with $[\beta_1, \beta_2] = [0.9, 0.99]$ and a batch size of 128. We provide an ablation study on the hyperparameter $\lambda$ of root loss $L_{root}$ in Tab.~\ref{tab:lambda_ablation}.
The latent dimension $d_{c}$ and hidden size are set to 16 and 1024, respectively.
The latent dimension significantly impacts the compression rate,
while the hidden size affects the model's capacity.
Both factors influence reconstruction and subsequent generation quality,
requiring a careful trade-off between compression efficiency and generative performance.
Ablation studies on the latent dimension and hidden size are provided in Tab.~\ref{tab:tae_ablation}.
To further improve the quality of the reconstructed motion, we add a linear layer after the embedded Gaussian distribution parameters as a latent adapter to
get a lower-dimensional and more compact latent space
for subsequent sampling, as proposed in \cite{lcmv2}.

For the Transformer inside the AR model, we use the architecture akin to LLaMA \cite{llama} with 12 layers,
12 attention heads and 768 hidden dimension.
The ablation for different scales of the Transformer is provided in Tab.~\ref{tab:transformer_ablation}.
Block size is set to 78 and we choose RoPE \cite{roformer} as the positional encoding.
For the diffusion head after Transformer, we use MLPs with 1792 hidden dimension and 9 layers.
The output vectors of the Transformer serve as the condition of denoising via AdaLN \cite{adaln}.
We adopt a cosine noise schedule with 50 steps for the DDPM \cite{DDPM}
denoising process following \cite{MDM}.
During training, the minimum and maximum length of motion sequences are set to 40 and 300 for both datasets.
We insert an additional reference end latent at the end of each motion sequence to indicate the stop of generation.
For Two-Forward strategy, a cosine scheduler is employed to control the ratio of replaced motion tokens, which can be formulated as:
$\gamma_{t} = \frac{1}{2}(1-\cos(\frac{\pi t}{T}))$,
where $t$ is current iteration step and $T$ is the total number of iterations.
When $t=0$, $\gamma_{t}=0$, indicating that no generated motion tokens in the first forward pass are replaced, thus relying on the ground-truth motion tokens only.
When $t=T$, $\gamma_{t}=1$, indicating that all generated motion tokens in the first forward pass are replaced, thus relying on the generated motion tokens only.
We use the same optimizer as the Causal TAE and a batch size of 256.
The initial learning rate is 1e-4 after 10K warmup iterations and decay to 0 for another 90K iterations using cosine learning rate scheduler.
Our experiments are conducted on A800 GPUs.

\begin{table}
    \centering
    \setlength{\tabcolsep}{15pt}
    \renewcommand{\arraystretch}{1.3}
    \begin{tabular}{lcc}
    \toprule
        $\lambda$ & FID $\downarrow$ & MPJPE $\downarrow$ \\
    \midrule
        5.0 & 0.696 & 25.2 \\
        6.0 & 0.684 & 24.8 \\
        7.0 & \textbf{0.661} & \textbf{22.9} \\
        8.0 & 0.682 & 24.2 \\
        9.0 & 0.704 & 26.8 \\
    \bottomrule
    \end{tabular}
    \caption{\textbf{Analysis of $\lambda$} on the HumanML3D \cite{T2M} test dataset.}
    \label{tab:lambda_ablation}
\end{table}

\begin{table*}
    \centering
    \begin{tabular}{lcccccccc}
    \toprule
    \multirow{2}{*}{Methods} & \multicolumn{2}{c}{Reconstruction} & \multicolumn{6}{c}{Generation} \\
    
    \cmidrule(lr){2-3} \cmidrule(lr){4-9}
    ~ & FID $\downarrow$ & MPJPE $\downarrow$ & FID $\downarrow$ & R@1 $\uparrow$ & R@2 $\uparrow$ & R@3 $\uparrow$ & MM-D $\downarrow$ & Div $\rightarrow$ \\
    \midrule
    Real motion & - & - & 0.002 & 0.702 & 0.864 & 0.914 & 15.151 & 27.492 \\
    \midrule
    (12,512)    & 8.862 & 38.5 & 21.078 & 0.600 & 0.759 & 0.827 & 17.143 & 27.456 \\
    (12,1024)  & 1.710 & 31.2 & 12.778 & 0.628 & 0.779 & 0.845 & 16.756 & 27.408 \\
    (12,1280) & 2.035 & 32.9 & 12.872 & 0.624 & 0.780 & 0.849 & 16.587 & 27.455 \\
    (12,1792)  & 1.563 & 28.3 & 11.916 & 0.628 & 0.782 & 0.850 & 16.468 & 27.461 \\
    (12,2048)  & 1.732 & 28.9 & 13.394 & 0.611 & 0.770 & 0.831 & 16.852 & 27.417 \\
    \midrule
    (14,512)  & 2.902 & 33.6 & 16.612 & 0.607 & 0.772 & 0.836 & 16.947 & 27.328 \\
    (14,1024)  & 0.838 & 27.5 & 11.933 & 0.627 & 0.778 & 0.840 & 16.593 & 27.443 \\
    (14,1280)  & 0.919 & 26.4 & 12.603 & 0.603 & 0.772 & 0.841 & 16.863 & 27.414 \\
    (14,1792)  & 0.732 & 24.8 & 11.828 & 0.628 & 0.776 & 0.848 & 16.652 & 27.122 \\
    (14,2048)  & 1.370 & 26.5 & 12.261 & 0.621 & 0.768 & 0.841 & 16.734 & 27.417 \\
    \midrule
    (16,512)  & 1.300 & 30.3 & 14.096 & 0.605 & 0.770 & 0.839 & 16.882 & 27.306 \\
    (16,1024) & 0.661 & 22.9 & \textbf{11.790} & \textbf{0.631} & \textbf{0.802} & \textbf{0.859} & \textbf{16.081} & 27.284 \\
    (16,1280)  & 1.087 & 25.0 & 12.975 & 0.598 & 0.761 & 0.831 & 17.002 & 27.403 \\
    (16,1792)  & 0.540 & 22.0 & 11.992 & 0.630 & 0.767 & 0.846 & 16.644 & 27.419 \\
    (16,2048)  & 1.547 & 26.2 & 12.778 & 0.604 & 0.755 & 0.824 & 16.897 & 27.306 \\
    \midrule
    (18,512)  & 2.043 & 27.7 & 19.150 & 0.553 & 0.701 & 0.775 & 17.776 & 27.345 \\
    (18,1024)  & 0.656 & 23.4 & 11.838 & 0.619 & 0.775 & 0.840 & 16.816 & 27.356 \\
    (18,1280)  & 0.820 & 23.1 & 11.815 & 0.629 & 0.801 & 0.847 & 16.816 & 27.461 \\
    (18,1792)  & 1.045 & 22.1 & 12.514 & 0.612 & 0.774 & 0.840 & 16.915 & 27.412 \\
    (18,2048)  & 0.595 & 21.5 & 11.803 & 0.613 & 0.801 & 0.832 & 17.004 & 27.451 \\
    \midrule
    (20,512)  & 0.531 & 24.5 & 12.247 & 0.613 & 0.765 & 0.832 & 16.920 & 27.277 \\
    (20,1024)  & \textbf{0.379} & \textbf{19.9} & 11.814 & 0.630 & 0.765 & 0.847 & 16.802 & 27.485 \\
    (20,1280)  & 0.429 & 20.1 & 16.465 & 0.557 & 0.705 & 0.774 & 17.680 & \textbf{27.490} \\
    (20,1792)  & 0.548 & 20.1 & 11.845 & 0.616 & 0.776 & 0.842 & 16.919 & 27.392 \\
    (20,2048)  & 0.690 & 20.7 & 11.910 & 0.625 & 0.782 & 0.844 & 16.785 & 27.346 \\
    \bottomrule
    \end{tabular}
    \caption{\textbf{Ablation Study of different Causal TAE architecture designs} on HumanML3D \cite{T2M} test set. Each generation model remains the same.
    MPJPE is measured in millimeters.
    (16, 1024) indicates the latent dimension and hidden size of the Causal TAE.}
    \label{tab:tae_ablation}
\end{table*}

\section{Causal TAE Architecture}
\label{app:ctae_architecture}
\begin{figure}[h]
    \centering
    \includegraphics[width=0.48\textwidth]{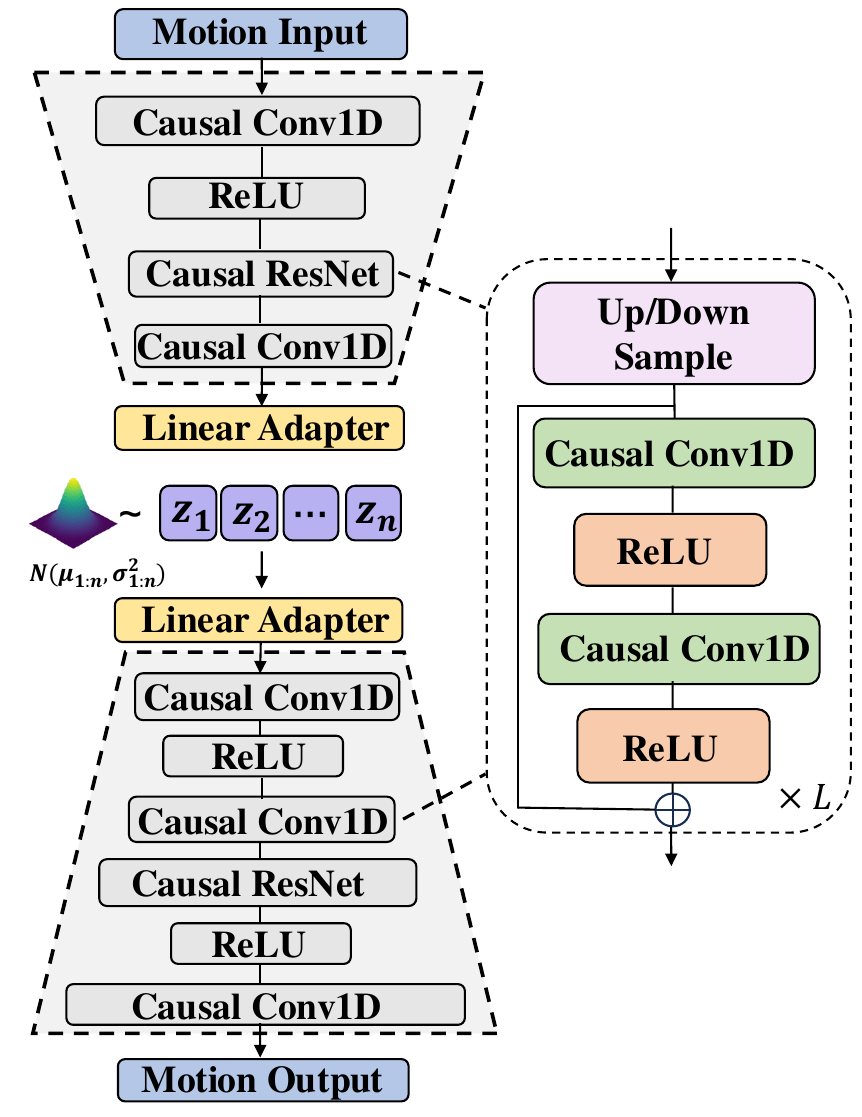}
    \caption{\textbf{Architecture of Causal TAE.} Motion latents are sampled in a continuous causal latent space.}
    \label{fig:vae_detail}
\end{figure}

The detailed architecture of the Causal TAE is shown in Fig.~\ref{fig:vae_detail} and Tab.~\ref{app:causal_vae_arch}.
Input motion sequences are first encoded into a latent space with a 1D causal ResNet.
The latent space is then projected to a sequence of Gaussian distribution parameters.
Then a linear adapter is applied to the embedded Gaussian distribution parameters to lower the dimension of latent space.
Sampling is performed in the lower-dimensional latent space.
The decoder comprises a mirror process to progressively reconstruct the motion sequence.

\section{AR Model Architecture}
We provide an ablation study on the architecture of the AR model, including the number of Transformer layers, attention heads, hidden dimension,
and the number of diffusion head layers, as shown in Tab.~\ref{tab:transformer_ablation}.
We finally leverage the 12-layer, 12-head, 768-hidden dimension, and 9-layer diffusion head architecture.

\section{Classifier-free guidance}
\begin{figure}
    \centering
    \includegraphics[width=0.48\textwidth]{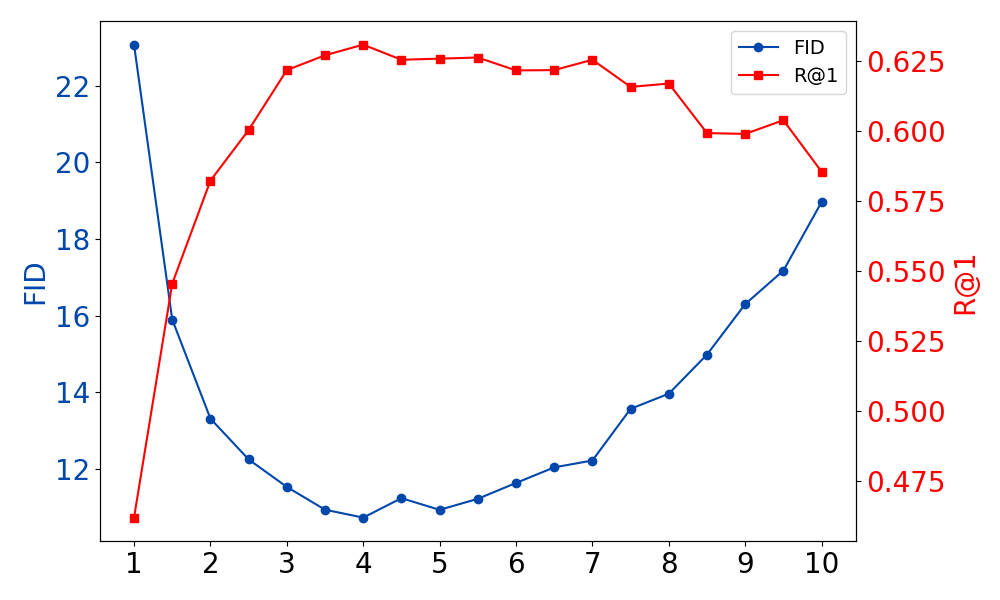}
    \caption{\textbf{Ablation of CFG scale} on HumanML3D \cite{T2M} test set.
    $scale=1$ means do not use CFG.}
    \label{fig:cfg}
\end{figure}

We adopt the classifier-free guidance (CFG) \cite{cfg} technique to improve the generation quality of the autoregressive motion generator.
Specifically, during training, we replace 10\% of the text within a batch with a blank text as unconditioned samples,
while during inference, CFG is applied to the denoising process of the diffusion head, which can be formulated as:
\begin{equation}
    \label{eq:cfg}
    \epsilon_{g} = \epsilon_{u} + s(\epsilon_{c}-\epsilon_{u}).
\end{equation}
where $\epsilon_{g}$ is the guided noise, $\epsilon_{u}$ is the unconditioned noise, $\epsilon_{c}$ is the conditioned noise,
$s$ is the CFG scale.
We provide an ablation study on the CFG scale $s$ in Fig.~\ref{fig:cfg}.
Finally, we choose $s=4.0$ for all experiments.

\begin{table*}[t]
    \centering\setlength{\tabcolsep}{12pt}
    \resizebox{1\textwidth}{!}{
    \begin{tabular}{cccccccccc}
        \toprule
        AR. layers & AR. heads & AR. dim & Diff. layers & FID $\downarrow$ & R@1 $\uparrow$ & R@2 $\uparrow$ & R@3 $\uparrow$ & MM-D $\downarrow$ & Div $\rightarrow$\\
        \midrule
        8 & 8 & 512 & 2 & 14.336 & 0.598 & 0.747 & 0.802 & 16.983 & 27.287\\
        8 & 8 & 512 & 3 & 13.764 & 0.602 & 0.758 & 0.819 & 16.972 & 27.242 \\
        8 & 8 & 512 & 4 & 12.893 & 0.608 & 0.764 & 0.828 & 16.661 & 27.351\\
        8 & 8 & 512 & 9 & 11.823 & 0.623 & 0.772 & 0.835 & 16.655 & 27.385\\
        8 & 8 & 512 & 16 & 12.460 & 0.621 & 0.778 & 0.849 & 16.784 & 27.410 \\
        12 & 12 & 768 & 2 & 11.899 & 0.601 & 0.763 & 0.828 & 16.952 & 27.406 \\
        12 & 12 & 768 & 3 & 11.798 & 0.630 & 0.779 & 0.844 & 16.761 & \textbf{27.482} \\
        12 & 12 & 768 & 4 & 12.051 & 0.604 & 0.762 & 0.829 & 16.940 & 27.401 \\
        12 & 12 & 768 & 9 & \textbf{11.790} & \textbf{0.631} & \textbf{0.802} & \textbf{0.859} & \textbf{16.081} & 27.284 \\
        12 & 12 & 768 & 16 & 11.825 & 0.624 & 0.773 & 0.844 & 16.757 & 27.341 \\
        16 & 16 & 1024 & 2 & 12.836 & 0.606 & 0.765 & 0.832 & 16.901 & 27.319 \\
        16 & 16 & 1024 & 3 & 12.436 & 0.601 & 0.761 & 0.830 & 16.919 & 27.302 \\
        16 & 16 & 1024 & 4 & 13.005 & 0.614 & 0.763 & 0.830 & 16.967 & 27.196\\
        16 & 16 & 1024 & 9 & 12.093 & 0.614 & 0.778 & 0.843 & 16.850 & 27.308 \\
        16 & 16 & 1024 & 16 & 11.812 & 0.630 & 0.780 & 0.846 & 16.598 & 27.286 \\
        \bottomrule
    \end{tabular}
    }
    \caption{\textbf{Ablation study of AR Model architecture} on HumanML3D \cite{T2M} test set.
    For each architecture, we use the same Causal TAE.}
    \label{tab:transformer_ablation}
\end{table*}

\section{Failure of Inverse Kinematics}
\label{app:ik}
\PAR{Post-processing for 263-dimensional motion representation.}
Most previous works \cite{MDM,mmask,T2M-GPT} uses 263-dimensional motion representation \cite{T2M}.

\begin{figure*}
    \centering
    \includegraphics[width=1\textwidth]{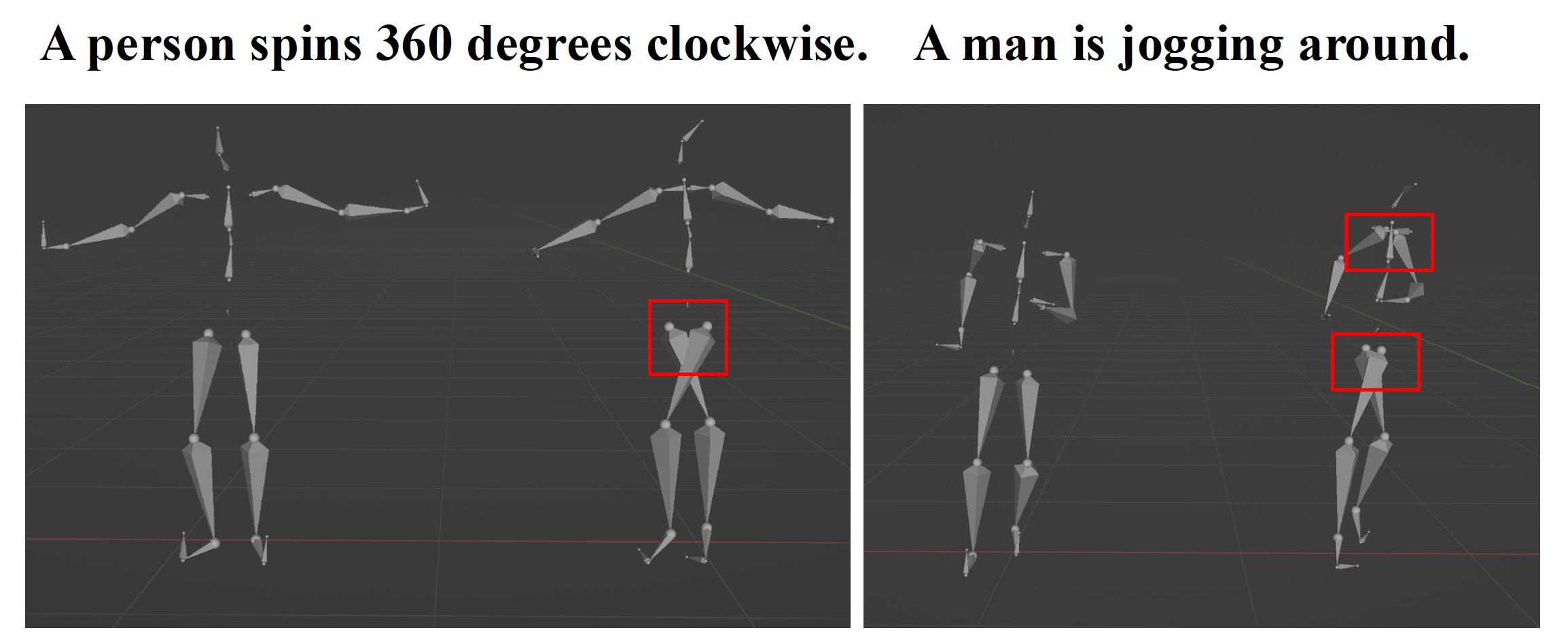}
    \caption{\textbf{Failure of Inverse Kinematics.} The joint rotation is directly solved using IK with relative joint positions,
    which leads to unnatural results like jittering body parts.}
    \label{fig:ik_failure}
\end{figure*}

The representation can be written as follows:
\setlength{\abovedisplayskip}{6pt} 
\setlength{\belowdisplayskip}{6pt} 
\begin{equation}
x = \{\dot{r}^a, \dot{r}^x,\dot{r}^z,r^y,\,j^p,\,j^v,\,j^r, c\},
\end{equation}
where the root is projected on the XZ-plane (ground plane),
\(\dot{r}^a \in \mathbb{R}^1\) denotes root angular velocity along the Y-axis,
\((\dot{r}^x, \dot{r}^z \in \mathbb{R})\) are root linear velocities on the XZ-plane,
\(r^y \in \mathbb{R}\) is the root height,
\(j^p \in \mathbb{R}^{3(K-1)}\), \(j^v \in \mathbb{R}^{3K}\), and \(j^r \in \mathbb{R}^{6(K-1)}\) are local joint positions, local velocities, and local rotations
relative to the root, \(K\) is the number of joints (including the root), and \(c\in \mathbb{R}^{4}\) is the contact label.
For SMPL characters, we have $K=22$ and we get $2 + 1 + 1 + 3\times21 + 3\times22 +6\times21 + 4 = 263$ dimensions.
In the original implementation \cite{T2M}, the joint rotation is directly solved using Inverse Kinematics (IK) with relative joint positions.
In such way, the joint loses twist rotation and directly applying the joint rotation to the character faces a lot of rotation error \cite{humanmlissue},
as shown in Fig.~\ref{fig:ik_failure}.
To overcome this issue, previous works \cite{MDM,mmask,T2M-GPT} only uses the positions and employs SMPLify \cite{Smplify} to solve the real SMPL joint rotation.
This process is time-consuming (around $60$ seconds for a $10$ seconds motion clip) and also introduces unnatural results like jittering head \cite{MDM}.
Most data in the HumanML3D \cite{T2M} dataset comes from the AMASS \cite{AMASS} dataset.
As the AMASS dataset provides the SMPL joint rotation, we slightly modify the motion representation by directly using the SMPL joint rotation and make it a 6D rotation for better learning.
Consequently, we remove the slow post-processing step and easily drive the SMPL character with the generated rotations.
The processing scripts to obtain our 272-dim motion representation are available at \url{https://github.com/Li-xingXiao/272-dim-Motion-Representation}.

\section{Limitations and Future work}
\PAR{Limitations.}
Despite its effectiveness, the streaming generation paradigm limits the applications of motion in-betweening and localized editing
of intermediate tokens, as it inherently relies on unidirectional modeling.
This limitation restricts flexibility in scenarios requiring fine-grained adjustments,
such as seamlessly inserting new motions between existing frames or interactively refining motion details while preserving global coherence.
\PAR{Future work.}
Future work could explore hybrid strategies that allow bidirectional refinement without compromising streaming generation.
One potential way is to predict a set of future latents at each step,
which could enable motion in-between and localized editing while preserving streaming manner.

\begin{table*}
    \centering\setlength{\tabcolsep}{6pt}
    \begin{tabular}{ll}
    \toprule
        Components & Architecture \\ \midrule
        Causal TAE Encoder & (0): CausalConv1D($D_{in}$, 1024, kernel\_size=(3,), stride=(1,), dilation=(1,), padding=(2,)) \\
        ~ & (1): ReLU() \\
        ~ & (2): 2 $\times$ Sequential( \\
        ~ &   ~~~~(0): CausalConv1D(1024, 1024, kernel\_size=(4,), stride=(2,), dilation=(1,), padding=(2,)) \\
        ~ &   ~~~~(1): CausalResnet1D( \\
        ~ &   ~~~~~~~~    (0): CausalResConv1DBlock( \\
        ~ &   ~~~~~~~~~~~~      (activation1): ReLU() \\
        ~ &   ~~~~~~~~~~~~      (conv1): CausalConv1D(1024, 1024, kernel\_size=(3,), stride=(1,), dilation=(9,), padding=(18,)) \\
        ~ &   ~~~~~~~~~~~~      (activation2): ReLU() \\
        ~ &   ~~~~~~~~~~~~      (conv2): CausalConv1D(1024, 1024, kernel\_size=(1,), stride=(1,), dilation=(1,), padding=(0,))) \\
        ~ &   ~~~~~~~~    (1): CausalResConv1DBlock( \\
        ~ &   ~~~~~~~~~~~~      (activation1): ReLU() \\
        ~ &   ~~~~~~~~~~~~      (conv1): CausalConv1D(1024, 1024, kernel\_size=(3,), stride=(1,), dilation=(3,), padding=(6,)) \\
        ~ &   ~~~~~~~~~~~~      (activation2): ReLU() \\
        ~ &   ~~~~~~~~~~~~      (conv2): CausalConv1D(1024, 1024, kernel\_size=(1,), stride=(1,), dilation=(1,), padding=(0,))) \\
        ~ &   ~~~~~~~~    (2): CausalResConv1DBlock( \\
        ~ &   ~~~~~~~~~~~~      (activation1): ReLU() \\
        ~ &   ~~~~~~~~~~~~      (conv1): CausalConv1D(1024, 1024, kernel\_size=(3,), stride=(1,), dilation=(1,), padding=(2,)) \\
        ~ &   ~~~~~~~~~~~~      (activation2): ReLU() \\
        ~ &   ~~~~~~~~~~~~      (conv2): CausalConv1D(1024, 1024, kernel\_size=(1,), stride=(1,), dilation=(1,), padding=(0,))))) \\
        ~ & (3): CausalConv1D(1024, 1024, kernel\_size=(3,), stride=(1,), dilation=(1,), padding=(2,)) \\
        \midrule
        Causal TAE Decoder & (0): CausalConv1D(1024, 1024, kernel\_size=(3,), stride=(1,), dilation=(1,), padding=(2,)) \\
        ~ & (1): ReLU() \\
        ~ & (2): 2 $\times$ Sequential( \\
        ~ &   ~~~~(0): CausalResnet1D( \\
        ~ &   ~~~~~~~~    (0): CausalResConv1DBlock( \\
        ~ &   ~~~~~~~~~~~~      (activation1): ReLU() \\
        ~ &   ~~~~~~~~~~~~      (conv1): CausalConv1D(1024, 1024, kernel\_size=(3,), stride=(1,), dilation=(9,), padding=(18,)) \\
        ~ &   ~~~~~~~~~~~~      (activation2): ReLU() \\
        ~ &   ~~~~~~~~~~~~      (conv2): CausalConv1D(1024, 1024, kernel\_size=(1,), stride=(1,), dilation=(1,), padding=(0,))) \\
        ~ &   ~~~~~~~~    (1): CausalResConv1DBlock( \\
        ~ &   ~~~~~~~~~~~~      (activation1): ReLU() \\
        ~ &   ~~~~~~~~~~~~      (conv1): CausalConv1D(1024, 1024, kernel\_size=(3,), stride=(1,), dilation=(3,), padding=(6,)) \\
        ~ &   ~~~~~~~~~~~~      (activation2): ReLU() \\
        ~ &   ~~~~~~~~~~~~      (conv2): CausalConv1D(1024, 1024, kernel\_size=(1,), stride=(1,), dilation=(1,), padding=(0,))) \\
        ~ &   ~~~~~~~~    (2): CausalResConv1DBlock( \\
        ~ &   ~~~~~~~~~~~~      (activation1): ReLU() \\
        ~ &   ~~~~~~~~~~~~      (conv1): CausalConv1D(1024, 1024, kernel\_size=(3,), stride=(1,), dilation=(1,), padding=(2,)) \\
        ~ &   ~~~~~~~~~~~~      (activation2): ReLU() \\
        ~ &   ~~~~~~~~~~~~      (conv2): CausalConv1D(1024, 1024, kernel\_size=(1,), stride=(1,), dilation=(1,), padding=(0,))))) \\
        ~ &   ~~~~(1): Upsample(scale\_factor=2.0, mode=nearest) \\
        ~ &   ~~~~(2): CausalConv1D(1024, 1024, kernel\_size=(3,), stride=(1,), dilation=(1,), padding=(2,)) \\
        ~ & (3) CausalConv1D(1024, 1024, kernel\_size=(3,), stride=(1,), dilation=(1,), padding=(2,)) \\
        ~ & (4): ReLU() \\
        ~ & (5): CausalConv1D(1024, $D_{in}$, kernel\_size=(3,), stride=(1,), dilation=(1,), padding=(2,)) \\
        \bottomrule
    \end{tabular}
    \caption{\textbf{Detail architecture} of the proposed Causal TAE.}
    \label{app:causal_vae_arch}
\end{table*}

\vspace{-250em}
\begin{figure}[h]
    \centering
    \includegraphics[width=0.29\textwidth]{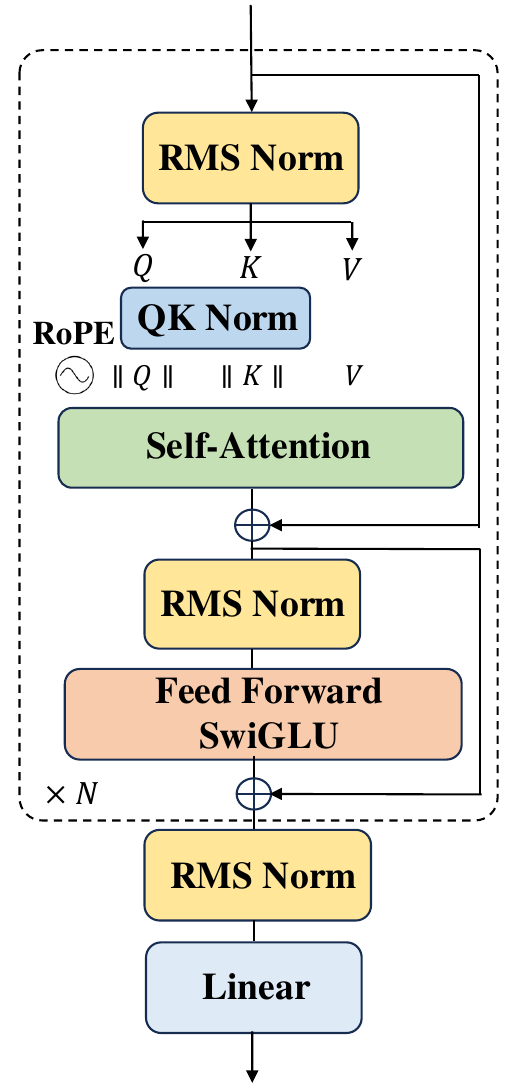}
    \caption{\textbf{Architecture of Transformer blocks in AR model.} QK Norm is applied to enhance training stability.}
    \label{fig:llama_detail}
\end{figure}


\end{document}